\documentclass[runningheads,a4paper]{llncs}

\let\subparagraph\relax 
\usepackage{titlesec}

\usepackage{amssymb}
\usepackage{graphicx}

\usepackage[T1]{fontenc}
\usepackage{lmodern}
\usepackage{microtype}
\usepackage{subfigure}
\usepackage{booktabs} 
\usepackage{amsmath}
\usepackage{bm}
\usepackage{authblk}
\usepackage{lineno}
\usepackage{float}
\usepackage[belowskip=-20pt,aboveskip=0pt]{caption}

\usepackage{url}
\urldef{\mailsa}\path|{astroseger, pas.aicv }@gmail.com |
\urldef{\mailsb}\path|{ hlatapie, efenogli}@cisco.com  |
\urldef{\mailsc}\path |    |       
\newcommand{\keywords}[1]{\par\addvspace\baselineskip
\noindent\keywordname\enspace\ignorespaces#1}

\begin{document}

\mainmatter  

\title{Metric Embedding Autoencoders for \\
Unsupervised Cross-Dataset Transfer Learning}

\titlerunning{Metric Embedding for Unsupervised Transfer Learning}

%
%
\ifx\anonym\undefined
\author{
	Alexey Potapov\inst{1,3} \and 
	Sergey Rodionov\inst{1,2} \and 	
	Hugo Latapie\inst{4} \and 
	Enzo Fenoglio\inst{4} }
\authorrunning{A. Potapov et al.}
\institute{  \textsuperscript{1}  SingularityNET Foundation\\
	\inst{2} Novamente LLC\\
	\inst{3} ITMO University, St. Petersburg, Russia\\
\mailsa\\
\inst{4} Chief Technology \& Architecture Office, Cisco \\
\mailsb }
\else
\author{Anonymous ICANN 2018 submission\inst{1}}
\authorrunning{Anonymized}
\institute{  \textsuperscript{1}  AnonymityNET}
 
\fi
%




%
%

\maketitle

\begin{abstract}
Cross-dataset  transfer learning is an important problem in person re-identification (Re-ID). Unfortunately, not too many deep transfer Re-ID models exist  for realistic settings of practical Re-ID systems. We propose a purely deep transfer Re-ID model consisting of a deep convolutional neural network and an autoencoder. The latent code is divided into metric embedding and nuisance variables. We then utilize an unsupervised training method that does not rely on co-training with non-deep models. Our experiments show improvements over both the baseline and   competitors' transfer learning models.
\end{abstract}

\section{Introduction}
\label{int}

Transfer learning is essential to most applications of deep learning in computer vision because of the scarcity of data available to train large networks in many tasks. The common practice is to take deep convolutional neural networks (DCNNs) such as ResNet-50 \cite{He15} or MobileNet \cite{Howard17} pre-trained on ImageNet \cite{Deng09} and fine-tune for the specific task by supervised learning on a subset of annotated samples. Actually, this practice can be considered as transferring features learned on a broad class of images from ImageNet to a more restricted domain.

\begin{figure}
\begin{center}
\centerline{\includegraphics[scale=0.75]{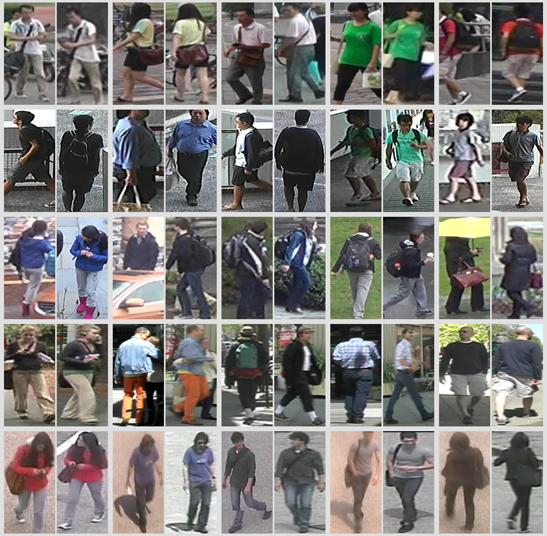}}
\caption{Pairs of images of same IDs from different cameras from different datasets: Market-1501 \cite{Zheng15}, CUHK03 ~\cite{Li14}, Duke ~\cite{Zheng17}, VIPeR ~\cite{Gray08}, WARD ~\cite{Martinel12}.}
\label{icml-fig1}
\end{center}
\end{figure}

However, it may be necessary to transfer a model, pre-trained via unsupervised learning, to a domain for which no labels are available. Person re-identification (Re-ID) can be considered as a motivating example as it consists of matching humans across cameras with non-overlapping fields of view. This task is challenging because of high variations in background, illumination, human poses, etc., and the absence of tight space-time constraints on candidate IDs such as in tracking, in addiction to re-identify persons absent in the training set. Even worse, it is usually necessary to deploy a person Re-ID system to a new camera set for which a large labeled training set is expensive or impossible to acquire, thus further motivating the use of pre-trained models for real-world applications. Unfortunately, if a model is trained on one dataset and tested on another, performances drop significantly below the level of hand-crafted features ~\cite{Fan17}, since variations between datasets are too large (see ~Figure~\ref{icml-fig1}).
For example, Rank-1 score can decrease from 0.762 to 0.361 on Market-1501 \cite{Zheng15} test set if the training is performed on DukeMTMC-reID \cite{Zheng17} training set instead of Market-1501 training set. Thus, it is essential to perform online unsupervised fine-tuning of pre-trained models. Generative models can provide a nice theoretical solution to the problem of unsupervised learning and transfer learning by constructing a generative model with the latent code containing different parts. The generative model can be fine-tuned in an unsupervised manner by marginalizing over unknown factors of variation.

State-of-the-art results in different tasks are usually achieved with discriminative models. Metric embedding learning ~\cite{Hermans17} or Siamese DCNNs \cite{Varior16} are successfully used in the Re-ID task, although without any capabilities of transferring to new camera sets. Generative models are not as \textit{deep} as discriminative models, and are not pre-trained on large datasets. Actually, they are  tested on simple domains, such as MNIST, \cite{Chen16}, \cite{Makhzani15} with limited practical applicability. Moreover, these models can utilize additional simplifications such as explicit one-hot coding of IDs \cite{Makhzani15}, which are not applicable in the Re-ID task. As a result, heuristic methods for unsupervised fine-tuning of the state-of-the-art discriminative models, such as the Progressive Unsupervised Learning (PUL) method applied to classification features ~\cite{Fan17}), are still beneficial.

\section{Related Work}
\label{rwx}
\subsection{Deep Re-ID methods}
Success of DCNNs in different applications of computer vision did not achieve acceptable performance on the task of person Re-ID, and a number of deep Re-ID models based on DCNNs have been proposed in recent years \cite{Li14},\cite{Yi14},\cite{Ahmed15}. The most popular approach is a deep metric learning with pairwise verification loss. In particular, Siamese DCNNs initially proposed in \cite{Yi14} for Re-ID are frequently used \cite{Ahmed15}, \cite{Shi16}, \cite{Varior16} for this purpose. This approach requires executing a model on each pair of a query and an image gallery. A more scalable approach is to learn a metric embedding (using triplet loss) which maps each image in the feature space where semantic similarity between images can be calculated using simple metrics.  A number of models has been developed, \cite{Khamis14}, \cite{Cheng16}, but cannot compare well with the models trained with classification and verification losses. However, Hermans et al. \cite{Hermans17} achieve  state-of-the-art results using metric embedding which we have also chosen for practical reasons. Many original Re-ID models exist, but they are out of scope since the focus of our work is on the problem of transferring models to new domains.
\subsection{Deep Transfer Learning for Re-ID}
There are different approaches to cross-dataset transfer learning for Re-ID. Some  utilize dictionary learning methods \cite{Peng16} and $\mathit{l}_1$ graph learning \cite{Kodirov16}, which are not deep. In these papers, the results are usually demonstrated on cases of transferring models to small datasets to show their advantages in comparison to deep learning models, which usually require large datasets. However, the work of Geng et al. \cite{Geng16} which uses co-training of a DCNN model and a graph regularised subspace learning model for unsupervised transfer learning, shows the potential to fine-tune DCNNs on the same small datasets (e.g. VIPeR ~\cite{Gray08} or PRID ~\cite{Hirzer11}) in order to achieve better performance. Real Re-ID systems can gather a large unlabeled amount of data quickly. A recent work by Fan et al. \cite{Fan17} describes a PUL method consisting of simultaneous improvement of the DCNN model and person clustering, and  conducted experimental validations on larger modern datasets including Market-1501 \cite{Zheng15} and Duke \cite{Zheng17}. We consider Fan et al. \cite{Fan17} more practical and realistic for our purposes, while assuming PUL a baseline for our comparison. The contributions of our work are as follows:
\begin{itemize}
\item A new, purely deep neural architecture is developed for cross-dataset transfer learning, consisting of a DCNN and an autoencoder, which latent code is divided into embedding and nuisance variables.
\item A method for training the proposed model is described, which preserves the properties of metric embedding during autoencoder unsupervised pre-training and fine-tuning.
\item Experiments are conducted showing considerable improvements over the baseline method.
\end{itemize}
\section{Metric Embedding Learning}
\label{meb}
\subsection{Loss Function}
For person Re-ID, it is usually assumed that bounding boxes (BBs) around humans are already extracted. BBs are usually resized to a  fixed size. Each BB yields a pattern (image) in an initial space of raw features ~${\mathbf{x}{\in}\mathbb{R}^N}$. BBs containing certain IDs can  be tracked by each camera, forming \textit{tracklets}, and in practice, it is better to compare tracklets instead of separate BBs. Each image ~$\mathbf{x}$ corresponds to a certain ID ~$\mathit{y}$, and the task is to identify which images from different cameras have the same ID. The IDs can be considered surrogate of classes, where the number of  classes is large and unknown while the number of images in each class is small. Therefore, it is inefficient to cast the Re-ID task as a traditional pattern recognition problem. One way to solve this problem is to train a model with a Siamese network that accepts two images as input and infers whether the two have the same ID. In this approach,  the model is run for one query image for each gallery image. Another option is to train a classification model with an DCNNs for a fixed set of IDs known for a training set,  cut off the fully connected (classification) layers, and compare images using high-level convolutional features which were useful for the classification. Similarity between images can be calculated directly as distance between  latent features with acceptable performance in the practical cases. However, in the non-linear space of features useful for classification, images with the same ID will not be necessarily closer together than images with different ID. An additional step of metric learning is mandatory to improve the overall performance.  Actually, what we want to learn is a metric embedding, i.e. a mapping ${f(\mathbf{x}|\bm{\theta})\colon \mathbb{R}^N{\rightarrow}\mathbb{R}^M}$ that transforms semantically similar images  onto metrically close points in $\mathbb{R}^M$, and semantically dissimilar images onto metrically distant points, i.e. $D_{i,j}{=}D(f(\mathbf{x}_i|\bm{\theta}), f(\mathbf{x}_j|\bm{\theta}))$ is small if $y_i{=}y_j$ and large otherwise, where $D$ is some metric distance measure (e.g. Euclidean \cite{Hermans17}). One can try to learn this mapping directly without learning the surrogate classification model, if an appropriate loss function is specified. In this case, the following triplet loss function can be used \cite{Hermans17}:

\begin{equation} \label{eq:1}
\mathcal{L}_{tri} {=} \sum_{\substack{a,p,n \\
                                 y_a = y_p \neq y_n}} 
                                 [m+D_{a,p}-D_{a,n}]_+
\end{equation}
where $m$ is some margin by which positive and negative examples should be separated. That is, different triplets of images are considered \textendash~ one is the anchor image with index $a$, the other is a positive example $y_p{=}y_a$ with index $p$, and the last one is a negative example $y_n{\neq} y_a$ with index $n$. We want the distance $D_{a,p}$ to be smaller than the distance $D_{a,n}$ by $m$. Softplus $ln(1+exp(x))$ is proposed in place of the hinge function $[m + \bullet]_+$ in ~\cite{Hermans17}, since in Re-ID we want to pull images with the same ID,  even after the margin $m$ is reached. Hard positive samples and hard negative samples shall be selected to make embedding learning with the triplet loss successful. Computationally efficient selection of  hard samples can be done with the use of \textit{Batch Hard} loss function \cite{Hermans17}. The idea is to form batches using $P$ randomly selected classes (IDs) with randomly sampled $K$ images per class, and to select the hardest positive and negative samples within the batch to form the triplets for the loss function  \cite{Hermans17}.

\subsection{Network Architecture}
We implemented the same network architecture as in \cite{Hermans17} with a few differences. Instead of ResNet-50, we used MobileNet \cite{Howard17}, since we found that the performance is very similar, while MobileNet is much faster. We also discarded the last classification layer and added two fully connected layers to map high-level convolutional features to the embedding space. Similarly, see Hermans et al. \cite{Hermans17}, we used the first dense layer with 1024 units with ReLU activation function, while the second (output) layer had 128 units corresponding to the embedding dimension. We also used batch normalization between layers.

\subsection{Embedding Training}
For the metric embedding training, we used ADAM optimizer with default parameters $(\beta_1{=}0.9, \beta_2{=}0.999)$. The learning rate was set to $10^{-4}$ during the first 100 epochs, and during the next 300 epochs we exponentially decayed the learning rate to $10^{-7}$. The number of steps per epoch was somewhat arbitrarily defined as $N_{total}/N_{batch}$, where $N_{total}$ is the total number of images in the datasets used, and $N_{batch}{=}K{*}P$ is the batch size. We used $K{=}4$ and $P{=}18$ in all experiments. We also applied embedding training on multiple datasets. Instead of simply merging the datasets together, we trained an embedding in such a way that the network never \textit{sees} images from different datasets simultaneously. We achieved this by forming each batch with images from only one dataset, and we continuously switched between them during training. This was done to prevent the model from simply pushing images from different datasets apart. Instead, this approach forced the model to search for invariant features, which will generalize to other datasets as well.

\section{Unsupervised Transfer Learning of Embedding}
\label{usle}
The problem with purely discriminative models  to transfer learning is that we do not have a criterion for fine-tuning unlabeled datasets. That is why a method such as PUL ~\cite{Fan17} uses a pre-trained model or some additional inputs to guess the reliability of positive and negative samples to use them with the same loss of supervised pre-training. To enable unsupervised transfer learning, we introduce a generative model describing the joint probability distribution $p(\mathbf{x}, \mathbf{z}_{id}, \mathbf{z}_{nui}, \mathbf{z}_{cam}| \bm{\theta})$, where $\mathbf{z}_{id}$ is the part of the latent code describing a specific person, $\mathbf{z}_{cam}$ describes a specific camera, $\mathbf{z}_{nui}$ is the vector of the rest nuisance variables (person pose and appearance, illumination conditions, etc.), and $\bm{\theta}$ is the parameter of the model. Since only few cameras are available and we do not have sufficient data to train this generative model,  we consider a model with camera-dependent parameters, i.e. $p(\mathbf{x}, \mathbf{z}_{id}, \mathbf{z}_{nui}| \bm{\theta}$(cam)). We want to train this model marginalizing over latent variables on several datasets to get the parameters $\bm{\theta}$(cam), which will be applicable (non-optimally) to different cameras, and then fine-tune (specialize) for a specific camera without labeled data. Using a generative model for Re-ID, we want  the latent code $\mathbf{z}_{id}$ for IDs to be a metric embedding. One option is to train a generative model, e.g. Adversarial Autoencoders (AAE) \cite{Makhzani15}, using an additional update for $\mathbf{z}_{id}$ with the triplet loss. Unfortunately, the quality of embedding drops because the updates for the adversarial and reconstruction losses spoil it. If we take the embedding trained independently, we will not know the corresponding priors $p(\mathbf{z}_{id})$, and we cannot directly use this embedding within a generative model. Therefore, we performed an unsupervised fine-tuning of the embedding without knowing or enforcing the corresponding priors $p(\mathbf{z}_{id})$ and $p(\mathbf{z}_{nui})$.

\subsection{Our Solution}
For practical considerations, we show improvements on the state-of-the-art Re-ID model. The first step of our method is to train the embedding model $\mathbf{z}_{id}=f_{emb}(\mathbf{x}|\bm{\theta}_{emb})$ as described in Section \ref{meb}. We supplement this mapping with the mappings $\mathbf{z}_{nui}{=}f_{nui}(\mathbf{x}|\bm{\theta}_{nui})$ and $\mathbf{x}{=}f_{dec}(\mathbf{z}_{id}, \mathbf{z}_{nui}|\bm{\theta}_{dec})$. Here, $(f_{emb}, f_{nui})$ is an encoder with the latent code consisting of two parts \textendash~ $\mathbf{z}_{id}$ and $\mathbf{z}_{nui}$, and $f_{dec}$ is a decoder constituting together an autoencoder. In the second step of our method we train the autoencoder using the same available labeled datasets, on which the embedding was trained. Here, weights $\bm{\theta}_{emb}$ are kept frozen, and $\bm{\theta}_{nui}$ and $\bm{\theta}_{dec}$ are optimized to minimize the reconstruction loss. This gives the pre-trained autoencoder, i.e. one part of the latent code to which corresponds the state-of-the-art embedding mapping. We will call this model EmbAE. However, the parameters of the autoencoder are not optimized for the target cameras, for which only unlabeled data is available. Thus, the third step should be the unsupervised fine-tuning. We can try to learn the parameters of all parts of the model, including $\bm{\theta}_{emb}$, $\bm{\theta}_{nui}$ and, $\bm{\theta}_{dec}$. Even such straightforward fine-tuning of the whole autoencoder improves scores of the model on new datasets, but it is not the best approach since nothing prevents $\mathbf{z}_{id}$ and $\mathbf{z}_{nui}$ from mixing within it. A layman approach to prevent this, is to freeze $\bm{\theta}_{nui}$ on the unsupervised fine-tuning step. This method works in practice even though  $\bm{\theta}_{nui}$ should also depend on the dataset. We call this model EmbAE-fix$\bm{\theta}_{nui}$. It is possible to prevent $\mathbf{z}_{id}$ and $\mathbf{z}_{nui}$ from mixing by optimizing $\bm{\theta}_{emb}$ and $\bm{\theta}_{nui}$ separately. We developed the following two-step fine-tuning procedure: first, discard pre-trained $\bm{\theta}_{nui}$ and optimize it with reconstruction loss with fixed $\bm{\theta}_{emb}$ and $\bm{\theta}_{dec}$. Second, we optimize $\bm{\theta}_{emb}$ and $\bm{\theta}_{dec}$ using fixed new $\bm{\theta}_{nui}$. We  call this model EmbAE-new$\bm{\theta}_{nui}$. It appears that the mapping parameters $\bm{\theta}_{nui}$ are considerably different for different datasets. We also considered a model which has its own mapping $f_{nui}(\mathbf{x}|\bm{\theta}_{nui})$ for each dataset. It is fine-tuned similarly to EmbAE-new$\bm{\theta}_{nui}$, but during pre-training on multiple datasets it also maintains different values of $\bm{\theta}_{nui}$ for each of them. However, this model is outperformed by the model, in which different $\bm{\theta}_{nui}$ is learned for each camera and each dataset. We call this model EmbAE-cam$\bm{\theta}_{nui}$. The method consists in the following steps:
\vspace{-5pt}
\begin{itemize}
\item Offline training of the embedding with the triplet loss on one or several labeled datasets.
\item Offline training of the EmbAE with common or individual (for each camera) encoder part $f_{nui}(\mathbf{x}|\bm{\theta}_{nui})$.
\item Unsupervised fine-tuning with frozen or re-trained $\bm{\theta}_{nui}$.
\end{itemize}

\subsection{Model Details}
\begin{figure}
\centering
\centerline{\includegraphics[scale=0.6]{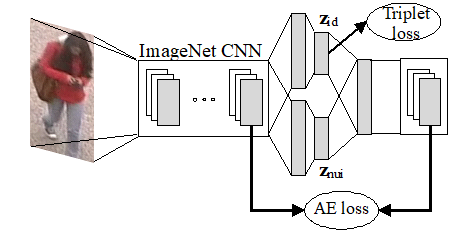}}
\caption{Deep Re-ID network architecture with unsupervised fine-tuning.}
\label{icml-fig2}
\end{figure}

The architecture for $f_{nui}(\mathbf{x}|\bm{\theta}_{nui})$ is the same as for $f_{emb}(\mathbf{x}|\bm{\theta}_{emb})$. Moreover, they share the same convolutional features of MobileNet. Only dense layers are independent, but with the same structure: dense layer with 1024 units and ReLU activations followed by batch normalization followed by dense layer with 128 units with linear activations. The decoder consists of the dense layer with 1024 units with ReLU activation followed by one more dense layer with the number of units corresponding to the number of highest-level convolutional features in MobileNet, the reconstruction loss is calculated for the MobileNet features. Our model network architecture (see Figure \ref{icml-fig2}) can be treated as an autoencoder with a truncated decoder, or in other words, that EmbAE is built on  top of MobileNet: it accepts convolutional features from MobileNet, and reconstructs these features \textendash~ not the original images.
\section{Experiments}
\label{exp}
We tested our approach using standard datasets CUHK03 ~\cite{Li14}, Duke ~\cite{Zheng17}, VIPeR ~\cite{Gray08}, WARD ~\cite{Martinel12} for training and Market-1501 ~\cite{Zheng15} for evaluation. Pre-training on a single dataset was used for comparison. Training on multiple datasets also helped to achieve higher scores. In our base architecture, we used one encoder for all images. In some cases, we used different encoders for images from different cameras of each dataset as described above. In all cases, we used only one embedding and one decoder.

\subsection{Score Computation}
To evaluate the models, we used  Rank-1 and mAP scores. For each image from the query set we searched the corresponding images in the test set. We let $I\!D_q$ and $C_q$ respectively be the image identity and the camera for a given query image $\mathit{q}$. Then, all images with $I\!D_q$ from camera $C_q$ are ignored and only images of $I\!D_q$ on cameras different from $C_q$ are assumed positive examples. Images with $I\!D_s$ different from $I\!D_q$ are assumed   negative samples, including images from camera $C_q$. In addition to the usual metrics, we consider scores calculated  ignoring all the images from camera $C_q$. We  refer to these scores as Rank-1-nd and mAP-nd. We use this score, because in real situations we will search only for images on other cameras, so negative examples from the same camera $C_q$ will not be considered. In our experiments, we used test-time data augmentation (see ~\cite{Hermans17} for details) in the score calculation. All networks were trained and tuned on data augmented by horizontal flip. We also used embedding normalization, i.e. we normalize $\mathbf{z}_{id}$ by its length: $\mathbf{z}_{id}/|\mathbf{z}_{id}|$ to increase the quality of models after unsupervised fine-tuning, because optimizing the reconstruction loss can distort the embedding space. The normalization was used only for score calculation.

\subsection{Single Dataset Pre-training}
Our first experiment was carried out for the models pre-trained on Duke dataset \cite{Zheng17}. Tests were performed on a different non-overlapping dataset, namely, Market-1501  \cite{Zheng15}. Table \ref{accuracy-table1} shows the results of the evaluation of different proposed architectures in comparison with the baseline model. The model with different encoders for different cameras provided the best results, and the improvement is rather large. EmbAE-new$\bm{\theta}_{nui}$ is no better than EmbAE-fix$\bm{\theta}_{nui}$. Thus, the better performance of EmbAE-cam$\bm{\theta}_{nui}$ is not simply  resulting from the optimization of $\bm{\theta}_{nui}$ for the specific dataset, but also to the increase of invariance of embedding w.r.t. cameras that helped to move all camera-variant features into $f_{nui}(\mathbf{x}|\bm{\theta}_{nui})$. 
\begin{table}
\vspace*{-20pt}
\caption{Re-ID accuracy of EmbAE trained on one dataset.}
\label{accuracy-table1}
\setlength\tabcolsep{2pt}
\centering
\scriptsize
\begin{sc}
\begin{tabular}{lclclclcl}
\toprule
Model & Rank-1 && Rank-1-nd && mAP && mAP-nd&\\
\midrule
Baseline    & 0.421 && 0.485 && 0.177 && 0.211 &\\\hline
EmbAE-fix$\bm{\theta}_{nui}$ & 0.553 &(+0.132)& 0.661 &(+0.176)& 0.275 &(+0.098)& 0.339 &(+0.128)\\\hline
EmbAE-new$\bm{\theta}_{nui}$ & 0.556 &(+0.135)& 0.650 &(+0.165)& 0.280 &(+0.103)& 0.337 &(+0.126)\\\hline
EmbAE-cam$\bm{\theta}_{nui}$ & 0.585 &(+0.164)& 0.669 &(+0.184)& 0.294 &(+0.117)& 0.345 &(+0.134)\\ 
\bottomrule
\end{tabular}
\end{sc}
\vspace*{-20pt}
\end{table}

\subsection{Multiple Dataset Pre-training}
We pre-trained our models using four datasets: Duke \cite{Zheng17}, CUHK03 \cite{Li14}, VIPeR \cite{Gray08}, WARD \cite{Martinel12}. Table \ref{accuracy-table2} shows the results of evaluation of these models in comparison with the baseline model.
\begin{table}
\vspace*{-20pt}
\caption{Re-ID accuracy of EmbAE trained on one dataset.}
\label{accuracy-table2}
\setlength\tabcolsep{2pt}
\centering
\scriptsize
\begin{sc}
\begin{tabular}{lclclclcl}
\toprule
Model & Rank-1 && Rank-1-nd && mAP && mAP-nd&\\
\midrule
Baseline    & 0.528 && 0.607 && 0.273 && 0.322 &\\\hline
EmbAE-fix$\bm{\theta}_{nui}$ & 0.596 &(+0.068)& 0.712 &(+0.105)& 0.329 &(+0.056)& 0.399 &(+0.077)\\\hline
EmbAE-new$\bm{\theta}_{nui}$ & 0.606 &(+0.078)& 0.707 &(+0.1)& 0.342 &(+0.069)& 0.404&(+0.082) \\\hline
EmbAE-cam$\bm{\theta}_{nui}$ & 0.643 &(+0.115)& 0.729 &(+0.122)& 0.357 &(+0.084)& 0.414 &(+0.092) \\ 
\bottomrule
\end{tabular}
\end{sc}
\vspace*{-15pt}
\end{table}
The model with different $\bm{\theta}_{nui}$ for each camera has the best scores. Although the improvements due to unsupervised fine-tuning became smaller, the final scores were much higher because the models properly pre-trained on several datasets were already considerably better.

\subsection{Comparison with PUL}
We are interested in training our model on a large high-quality dataset like Duke \cite{Zheng17} and also evaluating  on other large datasets. We compare the scores achieved by our model with PUL  method \cite{Fan17}, for which the results of transferring from both Duke and multiple datasets to Market-1501 \cite{Zheng15} are available. Table \ref{accuracy-table3} and Table \ref{accuracy-table4} show the results of this comparison, including the results obtained with the baseline models without transfer learning and improvements over these models from fine-tuning.
\begin{table}
\vspace*{-20pt}
\caption{Re-ID accuracy of PUL and EmbAE methods pre-trained on Duke.}
\label{accuracy-table3}
\setlength\tabcolsep{6pt}
\centering
\scriptsize
\begin{sc}
\begin{tabular}{lclcl}
\toprule
Model & Rank-1  & & mAP & \\
\midrule
Baseline PUL    & 0.361 & & 0.142 &  \\\hline
Fine-Tuned PUL & 0.447 & (+0.086) & 0.201&(+0.059) \\\hline
Baseline Embedding & 0.421 & & 0.273 &\\\hline
EmbAE-cam$\bm{\theta}_{nui}$ & 0.585 &(+0.164)& 0.294 & (+0.117)\\
\bottomrule
\end{tabular}
\end{sc}
\vspace*{ -10pt}
\end{table}
Despite that PUL uses an additional parameter (number of IDs in the new dataset), and that it was applied to improve the worse model, both the final scores and the improvements over the baseline models are better for our model, although still less than the models  trained in supervised manner and tested on Market-1501, which Rank-1 score can exceed 85\%.
\begin{table}
\vskip -20pt
\caption{Re-ID accuracy of PUL/EmbAE  pre-trained on multiple datasets.}
\label{accuracy-table4}
\setlength\tabcolsep{6pt}
\centering
\scriptsize
\begin{sc}
\begin{tabular}{lclcl}
\toprule
Model & Rank-1  & & mAP & \\
\midrule
Baseline PUL    & 0.400 & & 0.170 &  \\\hline
Fine-Tuned PUL & 0.455 &(+0.055) & 0.205 &(+0.035)\\\hline
Baseline Embedding & 0.528 & & 0.273 & \\\hline
EmbAE-cam$\bm{\theta}_{nui}$ & 0.643 &(+0.115)& 0.357 &(+0.084)\\
\bottomrule
\end{tabular}
\end{sc}
\vspace*{-20pt}
\end{table}
\section{Conclusion}
We have proposed a deep architecture for unsupervised cross-dataset transfer learning for person re-ID. This architecture is based on  metric embedding learning with  triplet loss function, which achieves state-of-the-art results \cite{Hermans17}. For transfer learning, metric embedding is incorporated into autoencoders. Special methods for pre-training and fine-tuning of autoencoders, which have a part of the latent code corresponding to metric embedding, have been proposed. These methods preserve embedding and prevent it from mixing with nuisance variables during unsupervised fine-tuning. Our experiments show improvements over  competitors' transfer learning models using the recent Progressive Unsupervised Learning method \cite{Fan17}, both in absolute scores and  over the baseline models.


\bibliographystyle{splncs04}
\bibliography{ICANN_article}
\end{document}